\documentclass[10pt,conference]{ieeeconf}      % Use this line for a4

\IEEEoverridecommandlockouts                              % This command is only
\usepackage{graphicx}          % Include this line if your
\usepackage{times,mathptmx}
\usepackage[T1]{fontenc}
\usepackage{color}
\usepackage{epsfig}
\usepackage[latin1]{inputenc}
\usepackage{rotating}
\usepackage{amsmath}
\usepackage{amssymb}
\usepackage{amsfonts}
\usepackage{epsf}
\usepackage{graphicx}
\usepackage{psfrag}
\usepackage{subfigure}
\usepackage{url}
\usepackage{tikz}
\usepackage{caption}
\usetikzlibrary{shapes,arrows}
\usepackage{pgfplots}
\usepackage{import}
\usepackage{algorithm}
\usepackage{comment}
\usepackage[noend]{algpseudocode}
\pgfplotsset{plot coordinates/math parser=false}

\renewcommand{\baselinestretch}{0.89}

\title{On Training and Evaluation of Neural Network Approaches for Model Predictive Control}

\author{Rebecka Winqvist, Arun Venkitaraman, Bo Wahlberg*% <-this % stops a space
\thanks{This work was partially supported by the Swedish Research Council and by the Wallenberg AI, Autonomous Systems and Software Program (WASP) funded by the Knut and Alice Wallenberg Foundation}% <-this % stops a space
\thanks{*Division of  Decision and Control Systems, School of Electrical Engineering and Computer Science, KTH Royal Institute of Technology, SE-100 44 Stockholm, Sweden.
 (e-mail: \{\texttt{rebwin, arunv, bo}\}@kth.se)}
}

\begin{document}

\maketitle
\thispagestyle{empty}
\pagestyle{empty}
\begin{abstract}   % Abstract of not more than 25
The contribution of this paper is a framework for training and evaluation of Model Predictive Control (MPC) implemented using constrained neural networks. Recent studies have proposed to use neural networks with differentiable convex optimization layers to implement model predictive controllers. The motivation is to replace real-time optimization in safety critical feedback control systems with learnt mappings in the form of neural networks with optimization layers. Such mappings take as the input the state vector and predict the control law as the output. The learning takes place using training data generated from off-line MPC simulations. However, a general framework for characterization of learning approaches in terms of both model validation and efficient training data generation is lacking in literature.
%
%The training data is generated by off-line simulations of the MPC. The aim of existing examples has been on demonstrating specific results while more systematic methods for data generation for training and  model validation  of such representations are lacking. 
In this paper, we take the first steps towards developing such a coherent framework. We discuss how the learning problem has similarities with system identification, in particular input design, model structure selection and model validation. We consider the study of neural network architectures in PyTorch with the explicit MPC constraints implemented as a differentiable optimization layer using CVXPY.
%involved in such CVXPY for the implementation of differentiable optimization layers with PyTorch for training of neural network based architectures. 
We propose an efficient approach of generating MPC input samples subject to the MPC model constraints using a hit-and-run sampler. The corresponding true outputs are generated by solving the MPC offline using OSOP. We propose different metrics to validate the resulting approaches. Our study further aims to explore the advantages of incorporating domain knowledge into the network structure from a training and evaluation perspective. Different model structures are numerically tested using the proposed framework in order to obtain more insights in the properties of  constrained neural networks based MPC.
%on a MPC benchmark from control of a heavy duty vehicle.

\end{abstract}

\section{Introduction}
\label{Sec1}

One of the challenges in the design of feedback control of safety critical systems is to ensure that the closed loop system always satisfies the given specifications. This is well understood when using classical control concepts as PID and LQR control. It is less is understood when it comes to constrained Model Predictive Control (MPC), which is usually implemented using on-line optimization. The lack of rigorous methods for verification of MPC often becomes a deterring factor against its widespread use in the vehicle industry despite its great potential. This in turn has prompted extensive research efforts in certified real-time optimization  for MPC over the recent years, see  \cite{9036084} for a overview. 

A common way to specify a MPC is through the use of a quadratic cost function with linear constraints, which handles both the dynamics, and the state and input  constraints. The optimal solution to a constrained quadratic program can be found by solving a set of linear equations once the active constraints are identified, see \cite{5153127} for results on fast  online MPC implementations. The main challenge lies in determining which of the constraints are active. Nevertheless, this entails solving an optimization problem at each time instant, which could quickly become a bottleneck when applying to systems repeatedly. One of the ways of circumventing the determination of active sets is through offline pre-computation of the control laws such that the problem is transformed into that of specifying a mapping or lookup table in the form of a piece-wise affine function. This mapping then acts on the input to produce the optimal control law. This approach is known as the explicit MPC, \cite{Alessio2009,10.5555/1965221}. The explicit MPC has further inspired the viewing of the MPC problem as a general learning problem. As a result, a number of works involving the use of learning approaches, primarily in the form of artificial neural networks have been proposed recently \cite{2018_paper,invariant_sets,chen2019large,maddalena2019neural}. The idea of using neural networks based MPC is by no means novel and there exist many publications on this topic. Early contributions include \cite{Parisini1995ARR,akesson2006}.
There many challenges with such learning based approaches, typically implemented using a neural network, both in terms of the modeling and the training and evaluation. There is a need for standardized framework for treatment of such approaches from an experimental design point of view. This forms the motivating force behind our current work.

%We will use OSQP, \cite{}, for solving QP. The underlying technique is based on ADMM combined with a strategy to identify active constraints using dual variables. Our experience is that OSQP is a very efficient and fast way to solve MPC problems for linear systems. 
%A natural idea then is to store pre-computed solutions to an off-line MPC optimization problem, and then use a look-up table to implement the control law for future problems. 
%There many challenges with such learning based approaches, typically implemented using a neural network. There is a need for standardized framework for treatment of such approaches from an experimental design point of view. This forms the motivating force behind our current work.
The main contributions of the paper is a framework for consistent analysis of learning based approaches for MPC in terms of the following key aspects:
\begin{itemize}
\item {\it Structure of the neural network}:  We study how to train a neural network that implements MPC using both the state and the model parameters as inputs, and the control law as the output of the neural network. We discuss different architectures, from the black box neural networks to architectures which are structure-aware in terms of the MPC problem.

\item {\it Nature of the data generation}: 
%A learning approach to MPC produces a control law which is a mapping from the measured state to the input control. 
It is not obvious how to efficiently generate training data when learning the control law as a mapping. This is related to input design for system identification \cite{7879927}. Grid-based approaches for sampling the input space would work for small state dimensions, but become cumbersome for high-dimensional systems. Keeping this in mind, we study the use of an efficient and statistically motivated hit-and-run sampler that extends well to higher state dimensions. We also discuss how gradient information can be used for input sample design and analyses. 
%In order to generate the corresponding training outputs of MPC solutions, we use the OSQP \cite{OSQP}, which is a state-of-the-art solver for quadratic optimization problems. 

\item {\it Evaluation of performance}: Different constrained neural networks approaches have recently been proposed in the literature for solving the MPC problem \cite{2018_paper,invariant_sets,chen2019large}. However, it generally remains unclear as to how these different structures can be evaluated consistently. Here, we study how different neural network approaches compare in terms of different metrics. In the process, we also investigate the importance of incorporating the MPC feasibility constraints in the form of a differentiable projection into the network. In other words, we study how structurally-aware network compares to a completely black box network.

%\item {\color{red}??}The complexity of the MPC mapping $u=\mu (x)$ depends on the constraint set.   We can obtain extra information by also calculating the gradient $d\mu (x)/dx$ . The training data generation should reflect this and  more examples are needed where the gradient  of $u=\mu (x)$ is large. 

%\item {\color{red}??}MPC is  based on a specific model and the corresponding control law is also a function of the model parameters.  We study how to train a neural networks implemented MPC with both the state and the model parameters as inputs.
\end{itemize}

The structure of this paper is as follows:   The MPC problem is introduced in Section \ref{Sec2}, while Section \ref{Sec3} describe the proposed training and evalution  framework. Network architectures for MPC are treated in  \ref{Sec4}, and  in Section  \ref{Sec5}  data generation and testing are studied. The numerical examples are presented in Section \ref{Sec6}. Finally, the conclusion and ideas for future work are given in Section \ref{Sec6}.

\section{Preliminaries}
\label{Sec2}
In this section, we briefly review the basic Model Predictive Control (MPC) problem, the explicit MPC, followed by a review of neural networks.

\subsection{Model Predictive Control}
Consider a discrete-time linear time-invariant system which evolves in time as
\begin{equation}
    \mathbf{x}_{k+1} = \mathbf{Ax}_k + \mathbf{Bu}_k,
    \label{eq:system}
\end{equation}
where $\mathbf{x}$ denotes the state vector and $\mathbf{u}$ denotes the input or control action, respectively; $\mathbf{A}$ and $\mathbf{B}$ denote the system matrices.
We will study the simplified MPC problem of steering the state of the system \eqref{eq:system} from an initial value to the origin by minimizing a control objective, subject to the state and input constraints
\begin{equation}
    \mathbf{x}_k \in \mathcal{X}, \quad \mathbf{u}_k \in \mathcal{U}
    \label{eq:constraints}
\end{equation}
where $\mathcal{X} \subseteq \mathbb{R}^n$ and $\mathcal{U} \subseteq \mathbb{R}^m$ are polyhedra representing the constraint sets for the state and the input, respectively.

%Consider now the problem of controlling \eqref{eq:system} to the origin or a fixed reference state, while subjected to the state and input constraints
%\begin{equation}
%    \mathbf{x}_k \in \mathcal{X}, \quad \mathbf{u}_k \in \mathcal{U}
%\end{equation}

 The MPC for an infinite time horizon entails solving the following constrained optimization problem
\begin{equation}
    \begin{aligned}
        & \min_{\bf{u}} & & \sum_{k=0}^{\infty}\mathbf{x}_k^T\mathbf{Q}\mathbf{x}_k
        + \mathbf{u}_k^T\mathbf{Ru}_k
        \\
        & \text{s.t.}
        & &\mathbf{x}_{k+1} = \mathbf{Ax}_k + \mathbf{Bu}_k, \\
        & {} & & \mathbf{x}_k \in \mathcal{X}, \\
        & {} & & \mathbf{u}_k \in \mathcal{U}, \\
        & {} & & \mathbf{x}_0 = \Bar{\mathbf{x}}
    \end{aligned}
    \label{eq:constrained_infinite_horizon}
\end{equation}
where $\mathbf{Q}$ and $\mathbf{R}$ are the positive semi-definite weight matrices. In an unconstrained setting of the state and input vectors, the solution to (\ref{eq:constrained_infinite_horizon}) results in the optimal feedback control law of the Linear Quadratic Regulator (LQR) given by
\begin{equation}
    \mathbf{u}_k^* =-\mathbf{Lx_k},\quad \mathbf{L}= (\mathbf{B}^T\mathbf{P}_{\infty}\mathbf{B} + \mathbf{R})^{-1}\mathbf{BP}_{\infty}\mathbf{A} 
\end{equation}
where $\mathbf{P}_{\infty}$ solves the \textit{Algebraic Riccati Equation}, \cite{mpc_book}.

In general the infinite time horizon MPC problem
 (\ref{eq:constrained_infinite_horizon}) is very difficult to solve, and one typically resorts to a finite-horizon MPC problem. Then, the following problem is solved at each instant
% \begin{comment}
% %{\color{red}[OSQP NOTATION]}
% \begin{align}
%     \begin{aligned}
%         & \min_{u} & & (\mathbf{x}_N-\mathbf{x}_r)^T\mathbf{Q}_N(\mathbf{x}_N-\mathbf{x}_r) + \sum_{k=0}^{N-1}(\mathbf{x}_k-\mathbf{x}_r)^T\mathbf{Q}(\mathbf{x}_k-\mathbf{x}_r) \\
%         & \text{s.t.}
%         & &\mathbf{x}_{k+1} = \mathbf{Ax}_k + \mathbf{Bu}_k, \\
%         & {} & &\mathbf{x}_{\text{min}} \leq \mathbf{x}_k \leq \mathbf{x}_{\text{max}}, \\
%         & {} & & \mathbf{u}_{\text{min}} \leq \mathbf{u}_k \leq \mathbf{u}_{\text{max}}, \\
%         & {} & & \mathbf{x}_0 = \Bar{\mathbf{x}}
%     \end{aligned}
% \end{align}
% \end{comment}
\begin{equation}
    \begin{aligned}
        & \min_{\mathbf{u}_0,\ldots, \mathbf{u}_{N-1}}& & J = \mathbf{x}_N^T\mathbf{Q}_N\mathbf{x}_N + \sum_{k=0}^{N-1}\mathbf{x}_k^T\mathbf{Q}\mathbf{x}_k
        + \mathbf{u}_k^T\mathbf{Ru}_k
        \\
        & \text{s.t.}
        & &\mathbf{x}_{k+1} = \mathbf{Ax}_k + \mathbf{Bu}_k, \\
        & {} & & \mathbf{x}_k \in \mathcal{X}, \\
        & {} & & \mathbf{u}_k \in \mathcal{U}, \\
        & {} & & \mathbf{x}_0 = \Bar{\mathbf{x}}
    \end{aligned}
    \label{eq:mpc}
\end{equation}
where $\mathbf{Q}_N$ is the terminal cost matrix and $N$ denotes the finite time horizon length. 

\begin{comment}
\begin{equation}
    \begin{aligned}
        & {\text{min}} & & (\mathbf{x}_N-\mathbf{x}_r)^T\mathbf{Q}_N(\mathbf{x}_N-\mathbf{x}_r) + \sum_{k=0}^{N-1}(\mathbf{x}_k-\mathbf{x}_r)^T\mathbf{Q}(\mathbf{x}_k-\mathbf{x}_r)
        + \mathbf{u}_k\mathbf{Ru}_k
        \\
        & \text{s.t.}
        & &\mathbf{x}_{k+1} = \mathbf{Ax}_k + \mathbf{Bu}_k, \\
        & {} & & \mathbf{x}_k \in \mathcal{X}, \\
        & {} & & \mathbf{u}_k \in \mathcal{U}, \\
        & {} & & \mathbf{x}_0 = \Bar{\mathbf{x}}
    \end{aligned}
\end{equation}
\end{comment}
\subsection{Feasibility}

\begin{figure}
    \centering
    \includegraphics[scale=0.4]{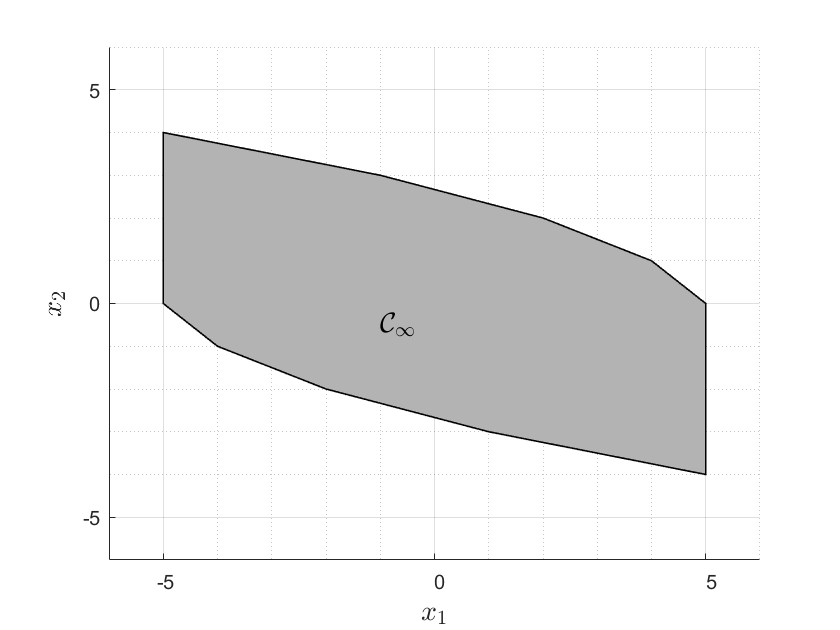}
    \caption{The maximal control invariant set $\mathcal{C}_{\infty}$ for system (\ref{eq:system}) with system and control matrices (\ref{eq:sys_con_matrices}) subject to the constraints (\ref{eq:constraints_def}).}
    \label{fig:exA_cinf}
\end{figure}

{
We now discuss how the feasibility constraints can be characterized in terms of set-invariance, which will form the basis of incorporating structure into neural network solutions for MPC.

Set invariance is closely connected with feasibility \cite{invariant_sets, mpc_book, kerrigan_2000}. From the definition in \cite{mpc_book}, a set, $\mathcal{C} \subseteq \mathcal{X}$, is a control invariant set for the system (\ref{eq:system}) subject to the constraints (\ref{eq:constraints}) if
\begin{equation}
\mathbf{x}_k \in \mathcal{C} \implies \exists \mathbf{u}_k \text{ s.t. } \mathbf{x}_{k+1} \in \mathcal{C}, \quad \forall k\in\mathbb{N}_+.
\end{equation}
In other words, for any initial state in $\mathcal{C}$ there exists a controller that ensures all future states reside in $\mathcal{C}$. The \textit{maximal control invariant} set, $\mathcal{C}_{\infty}$, is then defined as the control invariant set containing all control invariant sets contained in $\mathcal{X}$.

We shall see later how this set is used as a part of a projection strategy in neural networks for obtaining feasibility guarantees for the control law. To compute $\mathcal{C}_{\infty}$ we use algorithm 10.2 \textit{Computation of $\mathcal{C}_{\infty}$} provided in \cite{mpc_book} and the accompanying software. As already discussed in \cite{2018_paper}, we note that the algorithm comes with no termination guarantees but was found to converge in our experiments. 

In Figure \ref{fig:exA_cinf} we plot $\mathcal{C}_{\infty}$ for system (\ref{eq:system}) with 
\begin{equation}
        \mathbf{A} =
    \begin{bmatrix}
    1.0 & 1.0 \\
    0.0 & 1.0
    \end{bmatrix},
    \quad
    \mathbf{B} = 
    \begin{bmatrix}
    0.0 \\
    1.0
    \end{bmatrix}
    \label{eq:sys_con_matrices}
\end{equation}
}
subject to the constraints 
\begin{equation}
    \begin{bmatrix}
    -5.0 \\
    -5.0
    \end{bmatrix}
    \leq
    \mathbf{x}
    \leq
    \begin{bmatrix}
    5.0 \\
    5.0
    \end{bmatrix}, 
    \quad
    -2 \leq u_k \leq 2.
    \label{eq:constraints_def}
\end{equation}

{\subsection{Explicit MPC}}
{\color{black}
One considerable disadvantage of MPC is the computational effort required for solving (\ref{eq:mpc}) online, which makes MPC challenging for very  fast processes \cite{empc_bemporad}. Explicit MPC is a strategy for circumventing this issue by pre-computing the optimal control law offline. Given a polytopic set $\mathcal{X}$, for each $\mathbf{x} \in \mathcal{X}$ explicit MPC computes a piecewise affine (PWA) mapping from $\mathbf{x}$ to $\mathbf{u}$ defined over $M$ regions of $\mathcal{X}$.
% In this way, the explicit MPC performs a characterization of the control law akin to a look-up table: the only computation required online is then to determine what region the current state is in \cite{empc_bemporad, empc_survey}. While the explicit MPC on the one hand reduces the computational burden in terms of solving for different states, it typically does not scale well with the dimension of the state due to nature of the function \eqref{eq:xmpc}. {[\color{red} Q: Is the $\mathcal{X}$ here related to $\mathcal{C}_\infty$?]}

\begin{comment}
The computational effort required for solving (\ref{eq:mpc}) online is one of the main disadvantages/drawbacks of MPC, which makes it an unsuitable choice of control law in settings involving fast processes. 
\end{comment}

}

\subsection{Neural networks}
Neural networks and deep learning approaches are now ubiquituous and form the crux of most learning-based approaches today \cite{deeplearning}. Neural networks learn a mapping from the input to the output from known training examples, when the problem at hand has either no clear closed-form input-output mapping, or even if there is one, is intractable to work with \cite{Bishop}. Neural networks comprise concatenated processing units known as neurons that combine linear and non-linear transformations. Mathematically expressed, a neural network $n(\mathbf{x})$ learns a mapping from the input $\mathbf{x}$ to output $\mathbf{u}$ in the form
\begin{equation*}
    \mathbf{u}=n(\mathbf{x})=\sigma(\mathbf{W}_L \sigma(\mathbf{W}_{L-1}\sigma(\cdots \sigma(\mathbf{W}_1\mathbf{x}+\mathbf{b}_1)\cdots)+\mathbf{b}_L)
\end{equation*}
where $\sigma(\cdot)$ denotes the point-wise nonlinearity or activation function, and the matrices $\{\mathbf{W}_i,\mathbf{b}_i\}_{i=1}^L$ are the parameters (weights and biases) learnt by the network from the training data, $L$ denoting the number of neuron layers. The learning is typically performed by the use of back-propogation that uses the gradients of an error or loss function $n(\cdot)$ with respect to the network parameters. For the reasons of computational complexity and stability, the rectified linear unit (ReLU) is the most commonly employed activation function \cite{Bishop,deeplearning}. We note here that the use of ReLU as the activation function has been shown to be well motivated in the MPC setting due to its piece-wise linear nature \cite{2018_paper}. A schematic of a two-layer neural network is shown in Figure \ref{fig:noprojNN}.
\def\layersep{1cm}
\begin{figure}[t]
\centering
\begin{tikzpicture}[shorten >=1pt,->,draw=black, node distance=\layersep]
    \tikzstyle{every pin edge}=[<-,shorten <=1pt]
    \tikzstyle{neuron}=[draw,circle,fill=blue!60,minimum size=10pt,inner sep=0pt]
    \tikzstyle{input neuron}=[neuron, fill=gray!30];
    \tikzstyle{output neuron}=[neuron, fill=gray!30];
    \tikzstyle{hidden neuron}=[neuron, fill=gray!30];
    \tikzstyle{annot} = [text width=4em, text centered]
    \tikzstyle{missing} = [
    draw=none, 
    scale=1,
    text height=0.1cm,
    execute at begin node=\color{black}$\vdots$
  ]
  \tikzstyle{projection layer}=[draw, thick, rectangle, minimum height = 4em,
    minimum width = 3em, text centered, text width = 3.5em]
    \tikzstyle{fill neuron}=[circle,fill=white,minimum size=2pt,inner sep=0pt]

    % Draw the input layer nodes
    %\foreach \name / \y in {1,2,3}
    % This is the same as writing \foreach \name / \y in {1/1,2/2,3/3,4/4}
        %\node[input neuron, pin=left:$x_{\y}$] (I-\name) at (0,-\y) {};
    %\node[input neuron, pin=left:$\mathbf{x}$] (I-1) at (0,-1) {};
%    \node[missing] (m1) at (0,-2) {};
    \node[input neuron, pin=left:$\mathbf{x}_{k}$] (I-1) at (0,-2) {};
    %\node[missing] (m2) at (0,-4) {};
    %\node[input neuron, pin=left:$\mathbf{x}_{k+n}$] (I-3) at (0,-5) {};

    % Draw the hidden layer nodes
    \path[yshift=0.5cm]
    node[hidden neuron] (H-1) at (\layersep,-1 cm) {};
    \path[yshift=0.5cm]
    node[missing] (m3) at (\layersep,-1.7 cm) {};
    \path[yshift=0.5cm]
    node[hidden neuron] (H-2) at (\layersep,-2 cm) {};
    \path[yshift=0.5cm]
    node[hidden neuron] (H-3) at (\layersep,-3 cm) {};
    \path[yshift=0.5cm]
    node[missing] (m4) at (\layersep,-3.6 cm) {};
    \path[yshift=0.5cm]
    node[hidden neuron] (H-4) at (\layersep,-4 cm) {};
            
    % Draw the hidden layer nodes
    %\path[yshift=-1cm]
    %node[hidden neuron] (H2-1) at (2*\layersep,-1 cm) {};
    %\path[yshift=-1cm]
    %node[missing] (m5) at (2*\layersep,-2 cm) {};
    %\path[yshift=-1cm]
    %node[hidden neuron] (H2-2) at (2*\layersep,-3 cm) {};
    
    \path[yshift=0.5cm]
    node[hidden neuron] (H2-1) at (2*\layersep,-1 cm) {};
    \path[yshift=0.5cm]
    node[missing] (m5) at (2*\layersep,-1.7 cm) {};
    \path[yshift=0.5cm]
    node[hidden neuron] (H2-2) at (2*\layersep,-2 cm) {};
    \path[yshift=0.5cm]
    node[hidden neuron] (H2-3) at (2*\layersep,-3 cm) {};
    \path[yshift=0.5cm]
    node[missing] (m6) at (2*\layersep,-3.6 cm) {};
    \path[yshift=0.5cm]
    node[hidden neuron] (H2-4) at (2*\layersep,-4 cm) {};

    % Draw the output layer node
    %\node[output neuron,pin={[pin edge={->}]right:$y_1$}, right of=H2-3] (O1) {};
    
    %\node[output neuron,pin={[pin edge={->}]right:$\mathbf{u}_{k}^{P}$}] (O2) at (3*\layersep,-3 cm) {};
    \node[output neuron, pin={[pin edge={->}]right:$\mathbf{u}_{k}$}] (O2) at (3*\layersep,-2 cm) {};
    
    % Connect every node in the input layer with every node in the
    % hidden layer.
    \foreach \source in {1,...,1}
        \foreach \dest in {1,...,4}
            \path (I-\source) edge (H-\dest);
            
    \foreach \source in {1,...,4}
        \foreach \dest in {1,...,4}
            \path (H-\source) edge (H2-\dest);
    % Connect every node in the hidden layer with the output layer
    \foreach \source in {1,...,4}
        \path (H2-\source) edge (O2);

\end{tikzpicture}
\caption{Schematic of a two-layer neural network. We shall later refer to this architecture as the black box neural network (BBNN) since it does not actively use domain knowledge.}
\label{fig:noprojNN}
\end{figure}
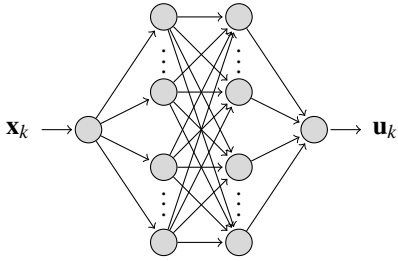

\section{Proposed framework}
\label{Sec3}
{\color{black}As discussed in the Section \ref{Sec1}, the goal of any learning approach to solving MPC problems is to obtain a meaningful mapping $\mu(\mathbf{x})$
such that for any given initial state $\mathbf{x}$, it produces the control law
\begin{equation*}
    \mathbf{u}=\mu(\mathbf{x}),
\end{equation*}
where $\mathbf{u}$ is the solution to the MPC. In order for one to make a consistent characterization of any learning approach, two important aspects have to be analyzed:
\begin{itemize}
    \item The nature of the mapping,
    \item The nature of data, in the context of training and evaluation.
\end{itemize}
Though learning approaches to MPC are not new, a framework for such a characterization of learning approaches in the MPC setting does not exist in literature.  
This motivates our present contribution in the form of a framework addressing these aspects, particularly in terms of training and evaluation through numerical experiments. We discuss these two aspects in detail next.

\subsection{The nature of $\mu(\mathbf{x})$}
Depending on the horizon, the state and input constraints, and the strategy employed, we have the following common variants of the MPC problem in general:
\begin{itemize}
    \item LQR, which corresponds to the unconstrained version of the infinite horizon problem \eqref{eq:constrained_infinite_horizon} ,
    
    \item Finite or infinite horizon control with constraints on the input and state vectors,
    
    \item Explicit MPC, which is a reformulation of the constrained MPC in terms of control invariant sets, giving a pre-computed offline lookup table-like characterization of the control law in terms of the state vector \cite{empc_bemporad},
    
    \item Learning based MPC approaches, which uses supervised learning from seen data examples typically in the form of a neural networks.
\end{itemize}
Correspondingly, each of these variants produce their own mapping $\mu(\mathbf{x})$. 
In the case of quadratic program based approaches for constrained MPC, $\mu(\mathbf{x})$ is obtained  by solving quadratic programs through convex optimization solvers such as the OSQP \cite{OSQP}. 

\subsection{The nature of data: Data generation for training and evaluation of $\mu(\mathbf{x})$}
Another key aspect in characterizing a learnt mapping is its relation to the available data-set. In terms of the training, this entails how the input space must be sampled such that the learning is meaningful and generalizes well. This aspect is of importance given that in many real-life applications such as vehicle control, one cannot be expected to have access to an unlimited number of diverse training data-points. 

In the case of the simple LQR,
$$
\mathbf{u}=-\mathbf{Lx}
$$
this poses no real difficulty since all that is needed to find $\mathbf{L}$ from training data is a single set of $n$ linearly independent samples of $x$ and corresponding $u$.

In the case of the explicit MPC, it comes with added complexity that one has to sample a set $n$ linearly independent points in each of the regions where the control law is constant.  For $q$ regions we thus need at least $n\times q$ samples of $\mathbf{x}$ and the corresponding $\mathbf{u}$.

Naturally, this issue also extends to the case where one wishes to work with neural network based learnt mappings -- how must we span the training space? One intuitively reasonable approach is that we uniformly sample the feasible set. To this end, we propose the use of a Hit-and-Run sampler which performs a random walk across the feasible set. In Figure \ref{fig:har_1000} we plot a set of 1000 points sampled using the hit-and-run sampler from the set $\mathcal{C}_{\infty}$ in Figure, \ref{fig:exA_cinf}. 
 The algorithm is presented in Section \ref{Sec5}.

We believe that this gives a consistent way of approaching the training aspect. Another related question, which we are currently pursuing, is whether it is more meaningful to incorporate the temporal aspect actively into the training process: rather than being randomly sampled isolated training data-points $\{\mathbf{x},\mathbf{u}\}$, should the training samples be from full trajectories?

The idea of representing a MPC with a structured neural network is also closely connected to the classical problem of function estimation from experimental data. This subject has a long history going back to the work of Box and co-workers on response surfaces \cite{Box}.  Gaussian processes and Bayesian optimization provide solid statistical frameworks for estimation of  function  and corresponding input design \cite{7352306}. Samples are typically chosen by optimizing a so-called acquisition function, which measures the current uncertainty of the function estimate. A simple method is  uncertainty sampling, where the next sample is taken where the uncertainty is the highest.  Unfortunately, less is understood regarding  optimal input design for the training of neural networks. %Since there is no explicit uncertainty in the data, we have a pure approximation problem.  
To obtain  insight into the approximation properties and generalization ability, we need tools to analyze the error and at the same time get information where to sample new data.
 Let $u=n(\mathbf{x})$ denote the network approximation of the MPC $\mathbf{u}=\mu(\mathbf{x})|$. On the training data we can calculate maximal error as
$\max_{\mathbf{x}_i\in\mathcal{D}}|n(\mathbf{x}_i)- \mu(\mathbf{x}_i)|$. 
Provided that the network is flexible enough these errors should be small after training. The challenge is to estimate the error for  $\mathbf{x}$ outside the sample set, and then sample additional  $\mathbf{x}$ where the errors are large. Recall that we also can calculate gradients with respect to $\mathbf{x}$ , i.e. $\nabla n(\mathbf{x})$ and  $\nabla d(\mathbf{x})$. Sub-gradients can be used  where the mappings are non-differentiable.
The gradient calculations can be done using CVXPY as described in \cite{agrawal2019differentiable}.
We then obtain information about the approximation error in a neighborhood around interesting sample points by computing the first order approximation
\begin{align*}
[n(\mathbf{x})- \mu(\mathbf{x})]&\approx  [n(\mathbf{x}_i)- \mu(\mathbf{x}_i)]\nonumber\\
&+ [\nabla n(\mathbf{x}_i)- \nabla n(\mathbf{x}_i)]^T(\mathbf{x}-\mathbf{x}_i)
\end{align*}
in a neighbourhood $||\mathbf{x}-\mathbf{x}_i||\leq \epsilon$ that does not overlap other existing samples of $\mathbf{x}$.
In order to generate new training data, we would like to identify regions where the approximation error is large. One way is to solve the QP
\begin{align*}
&\max_{\mathbf{x}}|| [\nabla n(\mathbf{x}_i)- \nabla d(\mathbf{x}_i)]^T(\mathbf{x}-\mathbf{x}_i)|| \nonumber\\ &\mbox{subject to}\;   ||\mathbf{x}-\mathbf{x}_i||\leq \epsilon
\end{align*}
for selected  $\mathbf{x}_i$ to obtain a measure where  to take new samples. 
 
Thus we see that it is possible to develop a well-motivated and systematic framework for the analysis and evaluation of learning based MPC. Our current contribution is a step in this direction.

\begin{figure}
    \centering
    \includegraphics[scale=0.4]{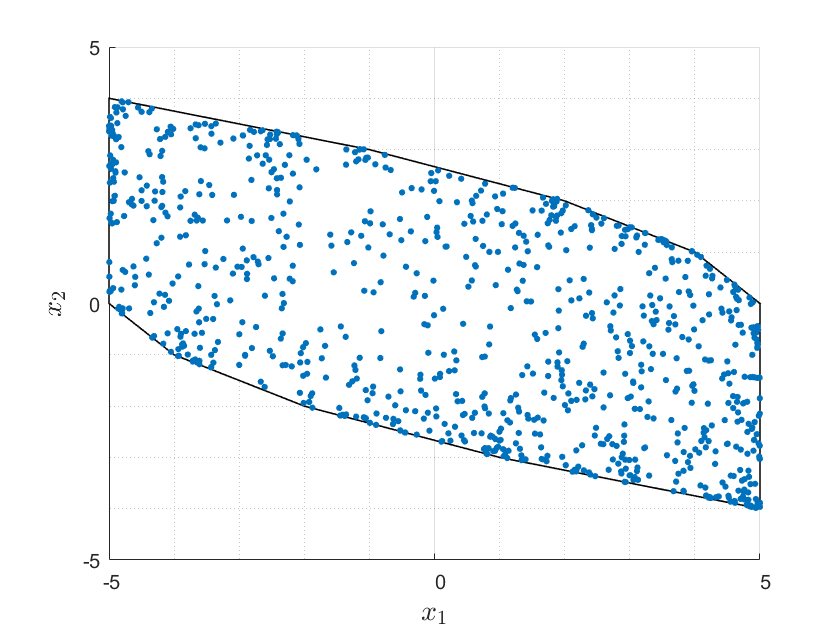}
    \caption{Hit-and-run sampling of 1000 points from the set $\mathcal{C}_{\infty}$ in Figure \ref{fig:exA_cinf}.}
    \label{fig:har_1000}
\end{figure}

}

\section{Network architecture}
\label{Sec4}
\import{}{network_architectures.tex}

\section{Training and data generation}
\label{Sec5}
We first discuss the systematic strategy that we propose for the generation of datasets for the MPC problems. Our approach for generating the training and testing data set involves sampling a set of states $\mathcal{S} = \{\mathbf{x}_1\, \hdots, \mathbf{x}_{N_S}\}$ from the feasible region $\mathcal{C}_{\infty}$ described in Section II-B, followed by solving for the optimal control inputs for each $s \in \mathcal{S}$ using OSQP \cite{OSQP}. To sample $\mathcal{C}_{\infty}$ we use the Hit-and-Run sampler which is a Markov chain Monte Carlo method for sampling uniformly from convex shapes \cite{METE20126}.  Essentially, starting from any point in the convex set, the method generates a set of points, $\mathcal{S}$, by walking random distances, $\lambda$, in randomly generated (unit) directions. The steps involved in the Hit-and-Run sampler are detailed in Algorithm \ref{alg:har}. We employ this approach since it ensures that the generated datapoints cover the feasibility region in a reasonably uniform manner \cite{METE20126, Zabinsky2013}. This then ensures that the network has observed training samples that span the entire feasible set on an average, thereby aiding its ability to generalize. Once the training and test inputs are generated, the corresponding training outputs are obtained from using OSQP on to solve the MPC problem.

\begin{algorithm}
\label{alg1}
\caption{Hit-and-Run Sampler}\label{alg:har}
\begin{algorithmic}[1]
\Procedure{Hit-and-Run }{$\mathcal{C}_{\infty}, N_S$}
\State Pick random point $\mathbf{x} \in \mathcal{C}_{\infty} = \{\mathbf{x} \in \mathbb{R}^n \mid \mathbf{C}_x\mathbf{x} \leq \mathbf{d}_x\}$
\State $\mathcal{S} \leftarrow \{\mathbf{x}\}$
\For{$i = 1, \hdots, N_S-1$}
\State $\lambda_i \leftarrow \infty$
\State Generate random unit direction $\mathbf{d}_i$
\For {$(\mathbf{c},d)$ in $(\mathbf{C}_x,\mathbf{d}_x)$}
\State $\lambda \leftarrow \displaystyle\frac{d-\mathbf{c}\cdot \mathbf{x}}{\mathbf{c}\cdot\mathbf{d}_i}$
\If {$\lambda > 0$} \Comment{To ensure right direction}
\State $\lambda_i \leftarrow \text{min}(\lambda_i, \lambda)$
\EndIf
\EndFor
\State $\lambda_i \leftarrow \text{drawn from }\mathbb{U}[0,\lambda_i)$
\State $\mathbf{x} \leftarrow \mathbf{x} + \lambda_i\mathbf{d}_i$
\State $\mathcal{S} \leftarrow \mathcal{S} \cup \{\mathbf{x}\}$
\EndFor
\State \textbf{return} $\mathcal{S}$
\EndProcedure
\end{algorithmic}
\end{algorithm}

We then use a supervised learning method to train the networks on a data set $\mathcal{D} = \{(\mathbf{x}_i,\mathbf{u}_i^*)\}$ of input-output pairs. 
During the training, we learn for the network parameters by minimizing a loss function $L(\theta)$ with respect to $\mathbf{\theta}$. 
%The most commonly used/default loss function for regression analysis is the Mean Square Error (MSE) \cite{mseloss}
In this work, we consider mean-square error as the loss function:
\begin{equation}
    L(\theta) = \frac{1}{N}\sum_{i=0}^{N-1}(n(\mathbf{x}_i,\mathbf{\theta}) - \mathbf{u}_i^*)^2.
\label{eq:mse_loss}
\end{equation}
%which is also what has been used in this work.
In order to increase the training speed, we split the data into smaller subsets (mini-batches) and compute the MSE-loss (\ref{eq:mse_loss}) for each batch. We then use the gradient descent-based \textit{Adam} optimizer \cite{adam} to backpropagate the loss and update the parameters $\mathbf{\theta}$ following each batch. Once all the mini-batches have been iterated over, one training epoch is completed. We train the networks for as many epochs required to reach convergence.

%\\\\
%For sampling $C_{\infty}$ we propose the Hit-and-Run sampler in Algorithm %\ref{alg:har}, a Markov chain Monte Carlo method for sampling uniformly from %convex shapes.
%\\\\
%For generating training and testing data sets, we propose a random walk approach %in which we start by sampling the feasible region $C_{\infty}$ hit-and-run sampler 
%\\\\
%To generate the training and testing data sets, we start by first sampling a %subset of $N_S$ states from the feasible region $C_{\infty}$ using the hit-and-run sampler in Algorithm \ref{alg:har}. We then proceed using OSQP \cite{osqp} to solve for the optimal control inputs for each state. 
%\\\\

\section{Examples}
\label{Sec6}
We now consider the application of the proposed concepts on three examples datasets. We evaluate the different network architectures in terms of two performance metrics:
\begin{enumerate}
    \item The normalized mean square error (NMSE) which is defined as
    \begin{align*}
        \mathrm{NMSE}=10\log_{10}\left(\frac{\|\mathbf{u}-\mathbf{u}^*\|_2^2}{\|\mathbf{u}^*\|_2^2}\right)
    \end{align*}
    \item The normalized control cost $J_n$, defined as the control cost, $J$, in Equation (\ref{eq:mpc}), normalized by $\mathbf{x}_0^T\mathbf{x}_0$:
    \begin{align*}
        J_n=  \frac{\mathbf{x}_N^T\mathbf{Q}_N\mathbf{x}_N+ \sum_{k=0}^{N-1}\left[\mathbf{x}_k^T\mathbf{Q}\mathbf{x}_k+\mathbf{u}_k^T\mathbf{R}\mathbf{u}_k\right]}{\mathbf{x}_0^T\mathbf{x}_0}
    \end{align*}
    where $\mathbf{x}_0$ is the initial state of the trajectory.
\end{enumerate}
Both the metrics are evaluated on test data, previously unseen by the networks during the training. NMSE helps evaluate the control law predicted by the network with respect to the ground truth, whereas the control cost measures how well the control law is in terms of minimizing the control objective: the smaller the $J$, the better the control achieved.

\subsection{Double integrator}
We first consider the example of a two-dimensional state vector with a scalar input under constraints. Despite being relatively low-dimensional, such a scenario occurs regularly in many real-life control applications. For example, in the case of a simplified bicycle model employed in autonomous vehicles in Scania \cite{BicycleScania2018}, the state model is given by
\begin{align*}
    x_{t+1}=\left[\begin{matrix}
    \cos(\Delta_t\kappa_t)&\sin(\Delta_t\kappa_t)\\
    -\kappa\cos(\Delta_t\kappa_t)& \sin(\Delta_t\kappa_t)
    \end{matrix}\right]x_t+\left[\begin{matrix}(1-\cos(\Delta_t\kappa_t)/\kappa_t^2\\
    \sin(\Delta_t\kappa_t)/\kappa_t\end{matrix}\right]u_t
\end{align*}
where the state vector consists of the direction coordinate and yaw, and $\Delta$ and $\kappa$ are parameters proportional to the velocity and curvature of the vehicle, respectively. The control objective matrices are usually of the form $\mathbf{Q}=\mathbf{I}$ and a scalar non-negative $\mathbf{R}$. Though the most general models are time-varying (as seen from the state matrices), it specializes to a time-invariant MPC problem when the velocity and curvature of the vehicle are kept constant (that is, when $\Delta_t$ and $\kappa_t$ are constant over time) \cite{BicycleScania2018}. This motivates us to consider first the case of a two-dimensional MPC problem.

Let us then consider a two-dimensional double integrator system specified as follows \cite{mpc_book}: 
\begin{equation}
    \mathbf{A} =
    \begin{bmatrix}
    1.0 & 1.0 \\
    0.0 & 1.0
    \end{bmatrix},
    \quad
    \mathbf{B} = 
    \begin{bmatrix}
    0.0 \\
    1.0
    \end{bmatrix}
    \label{eq:2D_system}
\end{equation}
subject to the constraints
\begin{equation}
    \begin{bmatrix}
    -5.0 \\
    -5.0
    \end{bmatrix}
    \leq
    \mathbf{x}_k
    \leq
    \begin{bmatrix}
    5.0 \\
    5.0
    \end{bmatrix},
    \quad
    -2.0 \leq u_k \leq 2.0,
    \quad k = 1, \hdots, N
    \label{eq:2D_constraints}
\end{equation}
and with cost parameters
\begin{equation}
    \mathbf{Q}_N = \mathbf{Q} = 
    \begin{bmatrix}
    1.0 & 0.0 \\
    0.0 & 1.0
    \end{bmatrix},
    \quad
    \mathbf{R} = 10,
    \quad
    N = 3.
    \label{eq:2D_costs}
\end{equation}
We generate training and testing data by sampling states from $\mathcal{C}_{\infty}$ for the system (\ref{eq:2D_system}) subject to (\ref{eq:2D_constraints}) using Algorithm 1, followed by solving for the optimal controls using OSQP with cost parameters (\ref{eq:2D_costs}).

\begin{figure}
    \centering
    \includegraphics[scale=0.65]{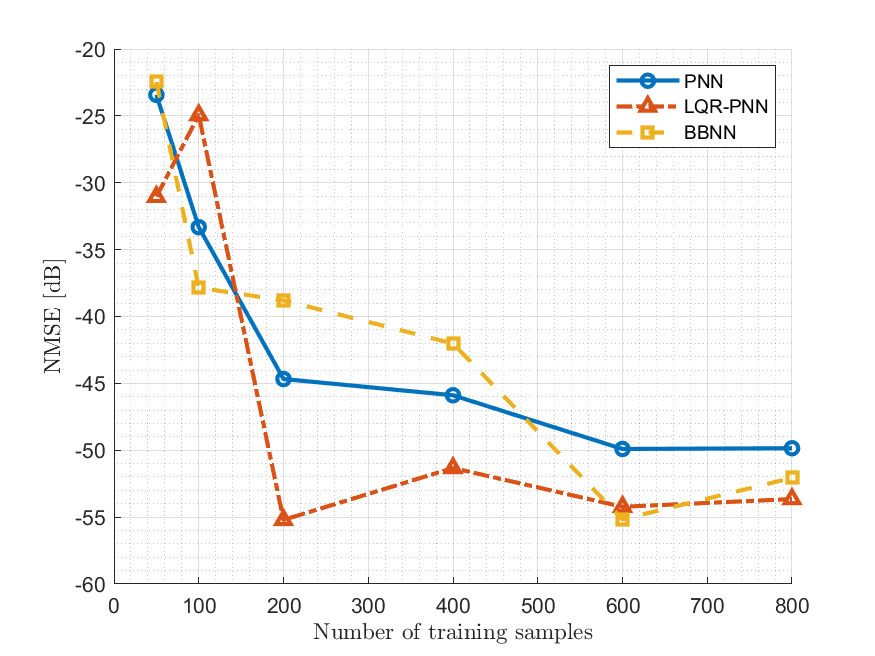}
    \caption{NMSE comparison for the test data 2D-example. }
    \label{fig:exA_lognmse}
\end{figure}

\begin{figure}
    \centering
    \includegraphics[scale=0.65]{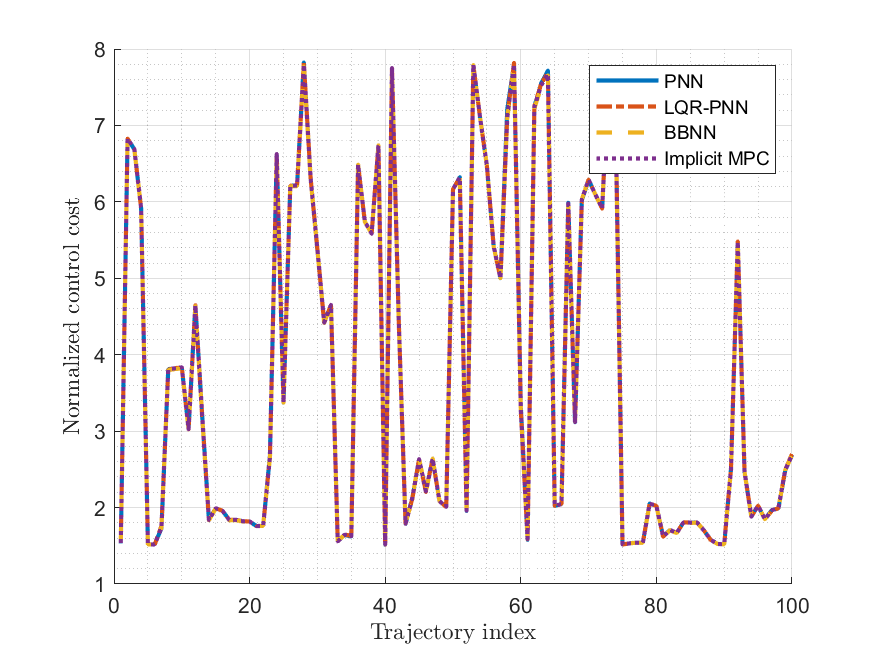}
    \caption{Comparison of the control costs computed for the different control laws for 100 test trajectories for the 2D-example.}
    \label{fig:exA_traj_100}
\end{figure}

In Figure \ref{fig:exA_lognmse}, we plot the NMSE as a function of the size of the training dataset. In Figure \ref{fig:exA_traj_100}, we show the normalized control cost, $J_n$, computed for 100 trajectories for the different control laws. For the NMSE evaluation, we use a testing dataset of 500 samples. For the control cost evaluation, we use a dataset of 1000 samples to train the networks. The initial states $\{\mathbf{x}_0^{(1)}, \hdots, \mathbf{x}_0^{(100)}\}$ of the trajectories are sampled from $\mathcal{C}_\infty$ using Algorithm 1. 

\subsection{4-Dimensional system}
In our second example we consider a 4-dimensional system from \cite{2018_paper}
\begin{equation}
    \mathbf{A} =
    \begin{bmatrix}
    0.7 & -0.1 & 0.0 & 0.0 \\
    0.2 & -0.5 & 0.1 & 0.0 \\
    0.0 & 0.1 & 0.1 & 0.0 \\
    0.5 & 0.0 & 0.5 & 0.5 \\
    \end{bmatrix},
    \quad
    \mathbf{B} = 
    \begin{bmatrix}
    0.0 & 0.1 \\
    0.1 & 1.0 \\
    0.1 & 0.0 \\
    0.0 & 0.0
    \end{bmatrix}
    \label{eq:4D_system}
\end{equation}
subject to the constraints
\begin{equation}
    \begin{bmatrix}
    -6.0 \\
    -6.0 \\
    -1.0 \\
    -0.5 
    \end{bmatrix}
    \leq
    \mathbf{x}_k
    \leq
    \begin{bmatrix}
    6.0 \\
    6.0 \\
    1.0 \\
    0.5
    \end{bmatrix},
    \quad
    \begin{bmatrix}
    -5.0 \\
    -5.0
    \end{bmatrix}
    \leq
    \mathbf{u}_k
    \leq
    \begin{bmatrix}
    5.0 \\
    5.0
    \end{bmatrix}
    \quad k = 1, \hdots, N
    \label{eq:4D_constraints}
\end{equation}
and with cost parameters
\begin{align}
    \mathbf{Q}_N = \mathbf{Q} = 
    \begin{bmatrix}
    1.0 & 0.0 & 0.0 & 0.0 \\
    0.0 & 1.0 & 0.0 & 0.0 \\
    0.0 & 0.0 & 1.0 & 0.0 \\
    0.0 & 0.0 & 0.0 & 1.0 
    \end{bmatrix},
    \quad
    \mathbf{R} =
    \begin{bmatrix}
    1.0 & 0.0 \\
    0.0 & 1.0 
    \end{bmatrix},
    \nonumber\\
    N = 10.
    \label{eq:4D_costs}
\end{align}
\begin{figure}
    \centering
    \includegraphics[scale=0.65]{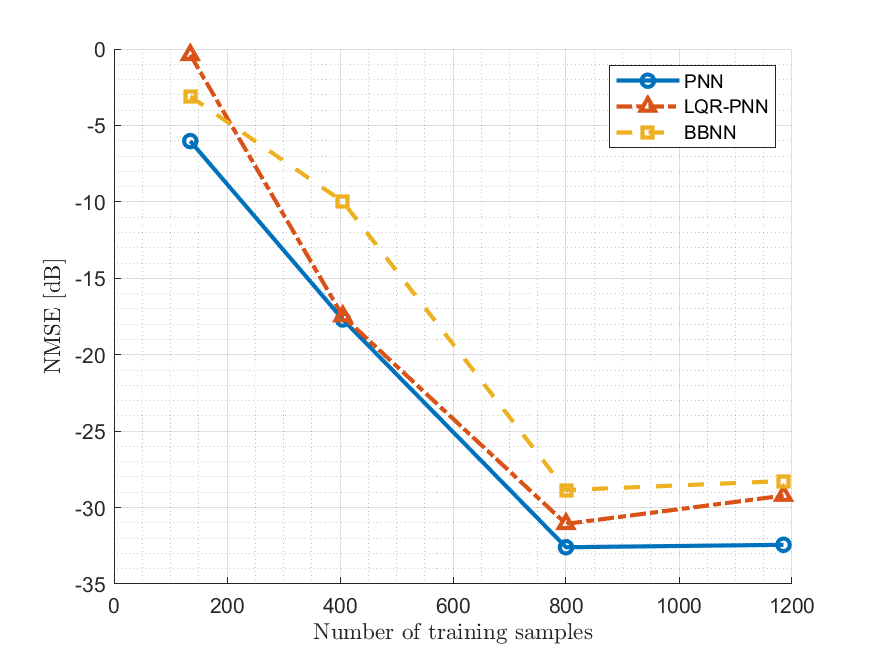}
    \caption{Comparison of the computed NMSE on test data for the different control laws for the 4D-example. }
    \label{fig:exE_lognmse}
\end{figure}

\begin{comment}
\begin{figure}
    \centering
    \includegraphics[scale=0.65]{plots/exE_traj_eval_70_plot.png}
    \caption{Comparison of the control costs computed for the different control laws for 70 test trajectories for the 4D-example.}
    \label{fig:exE_traj_70}
\end{figure}
\end{comment}

\begin{figure}
    \centering
    \includegraphics[scale=0.65]{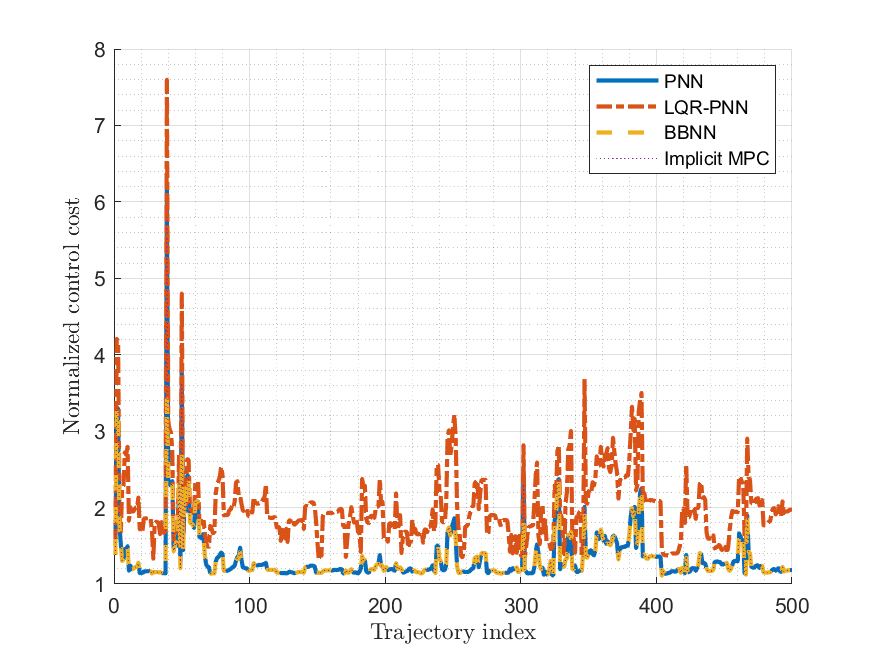}
    \caption{Comparison of the control costs computed for the different control laws for 500 test trajectories for the 4D-example.}
    \label{fig:exE_traj_500}
\end{figure}
We generate training and testing data by sampling states from $\mathcal{C}_{\infty}$ for the system (\ref{eq:4D_system}) subject to (\ref{eq:4D_constraints}) using Algorithm 1, followed by solving for the optimal controls using OSQP with cost parameters (\ref{eq:4D_costs}).

In Figure \ref{fig:exE_lognmse}, we plot the NMSE as a function of the size of the training dataset. In Figure \ref{fig:exE_traj_500}, we show the normalized control cost, $J_n$, computed for 500 trajectories for the different control laws. For the NMSE evaluation, we use a testing dataset of 500 samples. For the control cost evaluation, we use a dataset of 7000 samples to train the networks. The initial states $\{\mathbf{x}_0^{(1)}, \hdots, \mathbf{x}_0^{(500)}\}$ of the trajectories are sampled from $\mathcal{C}_\infty$ using Algorithm 1.

\section{Conclusion}
\label{Sec7}
We have presented a a framework for off-line training and evaluation of neural networks approaches for MPC. The basic idea is to approximate the MPC mapping from state to control input with constrained ReLU based neural networks including projection layer. Recent papers, \cite{invariant_sets,2018_paper} have motivated this structure from explicit MPC, and  that a continuous and piece-wise affine function on polyhedra can be represented exactly by a deep enough ReLu neural network. The role of the projection layer is to guarantee recursive feasibility and asymptotic stability. The main training and evaluation aspects we have studied are:
\begin{itemize}
\item We have used CVXPY \cite{agrawal2019differentiable} and PyTorch \cite{NEURIPS2019_9015} to implement this framework. We use OSQP \cite{OSQP} to generate training data and to evaluate the resulting controller. This tool box provides efficient means for training and evaluation. 
    \item A key factor is the generation of samples for the off-line training. This is a challenge when the dimension of the state space increase.  Here we proposed to use a hit and run sampler. We have also studied this problem from an experimental function approximation view. The generation of training data and evaluation is more computationally demanding than the training of the network.
    \item Evaluation of the resulting controller based on trajectories and  normalized cost-functions. The numerical tests shows the trade-off between  the number of training data and the approximation properties of the resulting controller.
\end{itemize}

This paper is to first step to more systematic methods for training and evaluation of neural networks implemented  MPC.  A major restriction is that current methods trains for a specific model. An interesting idea for future work is to use model parameters as well as states as input to the constrained network. This would increase the complexity and the need for even more structured training data generation.

The real controller use the measured state as input. Hence the training and evaluation should  take this into account. The good news is that less resolution in training data is needed, but the evaluation will become more challenging.  

The gradients of the MPC mapping and the gradient of the constrained network with respect to the state can be calculated. An interesting question is how this information can be used. Recall that the gain of the LQR controller
$
\mathbf{u}=-\mathbf{Lx}
$, can be found by just differentiating the control law at a given value of $\mathbf{x}$.

{\small

\bibliography{root.bbl}

\begin{thebibliography}{10}

\bibitem{agrawal2019differentiable}
Akshay Agrawal, Brandon Amos, Shane Barratt, Stephen Boyd, Steven Diamond, and
  Zico Kolter.
\newblock Differentiable convex optimization layers, 2019.

\bibitem{Alessio2009}
Alessandro Alessio and Alberto Bemporad.
\newblock {\em A Survey on Explicit Model Predictive Control}, pages 345--369.
\newblock Springer Berlin Heidelberg, Berlin, Heidelberg, 2009.

\bibitem{7879927}
M.~{Annergren}, C.~A. {Larsson}, H.~{Hjalmarsson}, X.~{Bombois}, and
  B.~{Wahlberg}.
\newblock Application-oriented input design in system identification: Optimal
  input design for control [applications of control].
\newblock {\em IEEE Control Systems Magazine}, 37(2):31--56, 2017.

\bibitem{empc_bemporad}
Alberto Bemporad.
\newblock {\em Explicit Model Predictive Control}, pages 1--9.
\newblock Springer London, London, 2013.

\bibitem{Bishop}
C.~M. Bishop.
\newblock {\em Pattern Recognition and Machine Learning (Information Science
  and Statistics)}.
\newblock Springer-Verlag New York, Inc., Secaucus, NJ, USA, 2006.

\bibitem{invariant_sets}
F.~Blanchini.
\newblock Set invariance in control.
\newblock {\em Automatica}, 35(11):1747 -- 1767, 1999.

\bibitem{mpc_book}
Francesco Borrelli, Alberto Bemporad, and Manfred Morari.
\newblock {\em Predictive Control for Linear and Hybrid Systems}.
\newblock Cambridge University Press, USA, 1st edition, 2017.

\bibitem{Box}
G.~E.~P. Box and K.B. Wilson.
\newblock On the experimental attainment of optimum conditions (with
  discussion).
\newblock {\em Journal of the Royal Statistical Society Series B}, 13(1):1--45,
  1951.

\bibitem{2018_paper}
S.~{Chen}, K.~{Saulnier}, N.~{Atanasov}, D.~D. {Lee}, V.~{Kumar}, G.~J.
  {Pappas}, and M.~{Morari}.
\newblock Approximating explicit model predictive control using constrained
  neural networks.
\newblock In {\em 2018 Annual American Control Conference (ACC)}, pages
  1520--1527, 2018.

\bibitem{chen2019large}
Steven~W. Chen, Tianyu Wang, Nikolay Atanasov, Vijay Kumar, and Manfred Morari.
\newblock Large scale model predictive control with neural networks and primal
  active sets, 2019.

\bibitem{9036084}
G.~{Cimini}, D.~{Bernardini}, S.~{Levijoki}, and A.~{Bemporad}.
\newblock Embedded model predictive control with certified real-time
  optimization for synchronous motors.
\newblock {\em IEEE Transactions on Control Systems Technology}, pages 1--8,
  2020.

\bibitem{cvxpy}
Steven Diamond and Stephen Boyd.
\newblock {CVXPY}: A {P}ython-embedded modeling language for convex
  optimization.
\newblock {\em Journal of Machine Learning Research}, 17(83):1--5, 2016.

\bibitem{10.5555/1965221}
Graham Goodwin, Mara~M. Seron, and Jos~A. de~Don.
\newblock {\em Constrained Control and Estimation: An Optimisation Approach}.
\newblock Springer Publishing Company, Incorporated, 1st edition, 2010.

\bibitem{kerrigan_2000}
Eric Kerrigan.
\newblock {\em Robust Constraint Satisfaction: Invariant Sets and Predictive
  Control}.
\newblock PhD thesis, Department of Engineering, University of Cambridge,
  Cambridge, 2000.

\bibitem{adam}
Diederik~P. Kingma and Jimmy Ba.
\newblock Adam: A method for stochastic optimization.
\newblock {\em CoRR}, abs/1412.6980, 2014.

\bibitem{deeplearning}
Y.~Bengio L.~Yann and G.~Hinton.
\newblock Deep learning.
\newblock {\em Nature}, 521(7553):436--444, 2015.

\bibitem{maddalena2019neural}
E.~T. Maddalena, C.~G. da~S.~Moraes, G.~Waltrich, and C.~N. Jones.
\newblock A neural network architecture to learn explicit mpc controllers from
  data, 2019.

\bibitem{METE20126}
Huseyin~Onur Mete and Zelda~B. Zabinsky.
\newblock Pattern hit-and-run for sampling efficiently on polytopes.
\newblock {\em Operations Research Letters}, 40(1):6 -- 11, 2012.

\bibitem{Parisini1995ARR}
Thomas Parisini and Riccardo Zoppoli.
\newblock A receding-horizon regulator for nonlinear systems and a neural
  approximation.
\newblock {\em Autom.}, 31:1443--1451, 1995.

\bibitem{NEURIPS2019_9015}
Adam Paszke, Sam Gross, Francisco Massa, Adam Lerer, James Bradbury, Gregory
  Chanan, Trevor Killeen, Zeming Lin, Natalia Gimelshein, Luca Antiga, Alban
  Desmaison, Andreas Kopf, Edward Yang, Zachary DeVito, Martin Raison, Alykhan
  Tejani, Sasank Chilamkurthy, Benoit Steiner, Lu~Fang, Junjie Bai, and Soumith
  Chintala.
\newblock Pytorch: An imperative style, high-performance deep learning library.
\newblock In H.~Wallach, H.~Larochelle, A.~Beygelzimer, F.~d\textquotesingle
  Alch\'{e}-Buc, E.~Fox, and R.~Garnett, editors, {\em Advances in Neural
  Information Processing Systems 32}, pages 8024--8035. Curran Associates,
  Inc., 2019.

\bibitem{BicycleScania2018}
G.~C. {Pereira}, P.~F. {Lima}, B.~{Wahlberg}, H.~{Pettersson}, and
  J.~{Mårtensson}.
\newblock Linear time-varying robust model predictive control for discrete-time
  nonlinear systems*.
\newblock In {\em 2018 IEEE Conference on Decision and Control (CDC)}, pages
  2659--2666, 2018.

\bibitem{7352306}
B.~{Shahriari}, K.~{Swersky}, Z.~{Wang}, R.~P. {Adams}, and N.~{de Freitas}.
\newblock Taking the human out of the loop: A review of bayesian optimization.
\newblock {\em Proceedings of the IEEE}, 104(1):148--175, 2016.

\bibitem{OSQP}
Bartolomeo Stellato, Goran Banjac, Paul Goulart, Alberto Bemporad, and Stephen
  Boyd.
\newblock Osqp: an operator splitting solver for quadratic programs.
\newblock {\em Mathematical Programming Computation}, pages 1867--2957, 2020.

\bibitem{5153127}
Y.~{Wang} and S.~{Boyd}.
\newblock Fast model predictive control using online optimization.
\newblock {\em IEEE Transactions on Control Systems Technology},
  18(2):267--278, 2010.

\bibitem{Zabinsky2013}
Zelda~B. Zabinsky and Robert~L. Smith.
\newblock {\em Hit-and-Run Methods}, pages 721--729.
\newblock Springer US, Boston, MA, 2013.

\bibitem{akesson2006}
Bernt Åkesson and Hannu Toivonen.
\newblock A neural network model predictive controller.
\newblock {\em Journal of Process Control}, 16:937--946, 10 2006.

\end{thebibliography}

}

\end{document}